\documentclass[letterpaper]{article} % DO NOT CHANGE THIS
\usepackage{aaai2026}  % DO NOT CHANGE THIS
\usepackage{times}  % DO NOT CHANGE THIS
\usepackage{helvet}  % DO NOT CHANGE THIS
\usepackage{courier}  % DO NOT CHANGE THIS
\usepackage[hyphens]{url}  % DO NOT CHANGE THIS
\usepackage{graphicx} % DO NOT CHANGE THIS
\usepackage{natbib}  % DO NOT CHANGE THIS AND DO NOT ADD ANY OPTIONS TO IT
\usepackage{caption} % DO NOT CHANGE THIS AND DO NOT ADD ANY OPTIONS TO IT
\usepackage{algorithm}
\usepackage{algorithmic}
\usepackage{amsmath}
\usepackage{amsfonts}
\usepackage{multirow}

\usepackage{hyperref}

\usepackage{newfloat}
\usepackage{listings}
\DeclareCaptionStyle{ruled}{labelfont=normalfont,labelsep=colon,strut=off} % DO NOT CHANGE THIS
\floatstyle{ruled}
\newfloat{listing}{tb}{lst}{}
\floatname{listing}{Listing}
\makeatletter
\g@addto@macro\@maketitle{%
  \vspace{0.5em}%
  \begin{center}
    \includegraphics[width=0.8\textwidth]{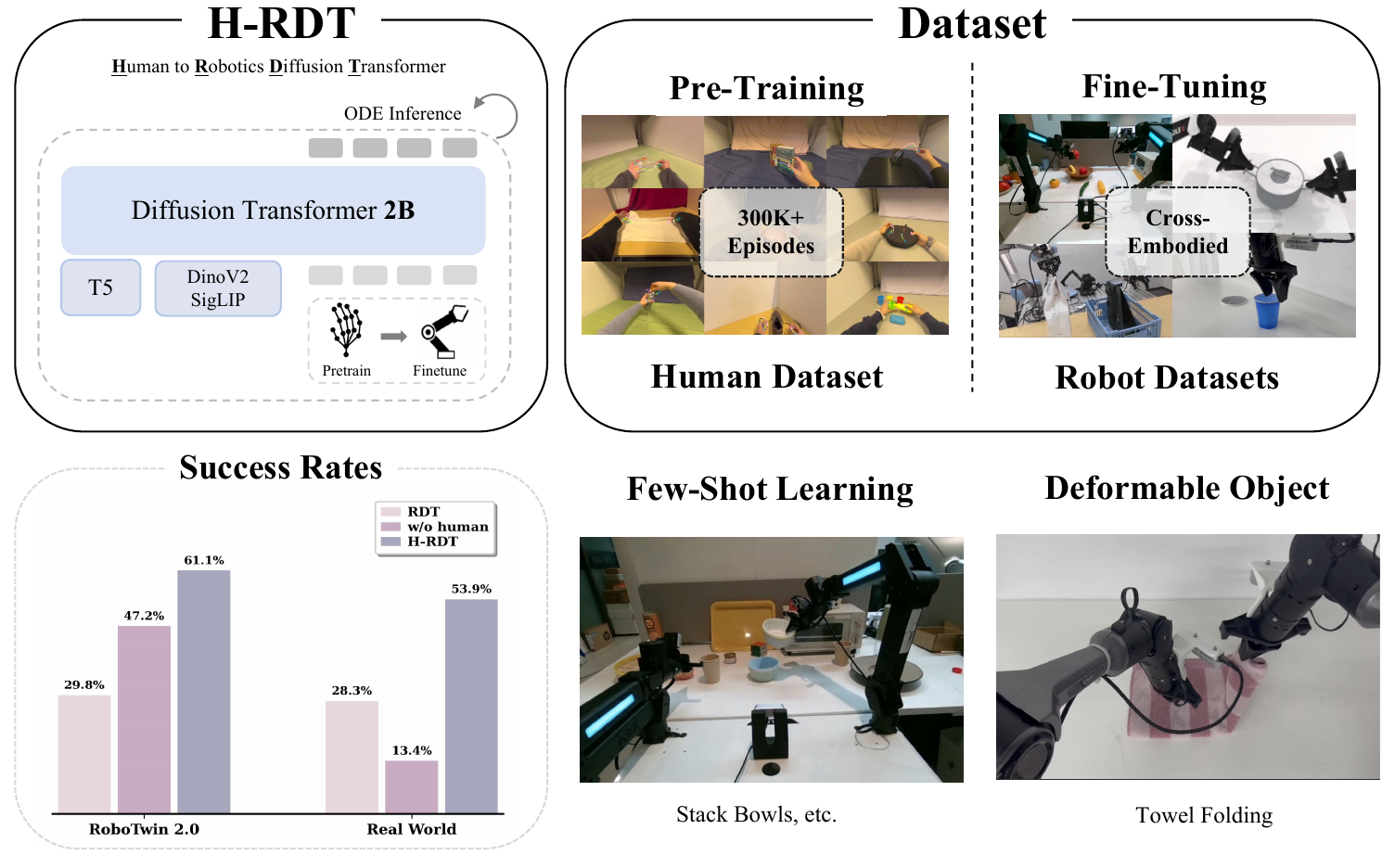}
    \captionof{figure}{\textbf{Overview of H-RDT.} A human-to-robotics diffusion transformer with two-stage training.}%
    \label{fig:performance_overview}
  \end{center}
  \vspace{0.8em}%
}%
\makeatother
\usepackage{xcolor}

\nocopyright

\title{H-RDT: Human Manipulation Enhanced Bimanual Robotic Manipulation}
\author{
    Hongzhe Bi\textsuperscript{\rm 1,\rm 2},
    Lingxuan Wu\textsuperscript{\rm 1},
    Tianwei Lin\textsuperscript{\rm 2},
    Hengkai Tan\textsuperscript{\rm 1},  \\
    Zhizhong Su\textsuperscript{\rm 2},
    Hang Su\textsuperscript{\rm 1},
    Jun Zhu\textsuperscript{\rm 1}
}
\affiliations{
    \textsuperscript{\rm 1}Dept. of Comp. Sci. and Tech., Institute for AI, BNRist Center, THBI Lab,\\
    Tsinghua-Bosch Joint ML Center, Tsinghua University\\
    \textsuperscript{\rm 2}Horizon Robotics\\
    bhz24@mails.tsinghua.edu.cn
}

\begin{document}

\maketitle

% Figure 1: Performance Overview (double-column, placed at top of first page)

\begin{abstract}
Imitation learning for robotic manipulation faces a fundamental challenge: the scarcity of large-scale, high-quality robot demonstration data. Recent robotic foundation models often pre-train on cross-embodiment robot datasets to increase data scale, while they face significant limitations as the diverse morphologies and action spaces across different robot embodiments make unified training challenging. In this paper, we present \textbf{H-RDT} (\textbf{H}uman to \textbf{R}obotics \textbf{D}iffusion \textbf{T}ransformer), a novel approach that leverages human manipulation data to enhance robot manipulation capabilities. Our key insight is that large-scale egocentric human manipulation videos with paired 3D hand pose annotations provide rich behavioral priors that capture natural manipulation strategies and can benefit robotic policy learning. We introduce a two-stage training paradigm: (1) pre-training on large-scale egocentric human manipulation data, and (2) cross-embodiment fine-tuning on robot-specific data with modular action encoders and decoders. Built on a diffusion transformer architecture with 2B parameters, H-RDT uses flow matching to model complex action distributions. The modular design of action encoder and decoder components enables effective knowledge transfer from the unified human embodiment to diverse robot platforms through efficient fine-tuning. Extensive evaluations encompassing both simulation and real-world experiments, single-task and multitask scenarios, as well as few-shot learning and robustness assessments, demonstrate that H-RDT outperforms training from scratch and existing state-of-the-art methods, including $\boldsymbol{\pi}_0$ and RDT, achieving significant improvements of \textbf{13.9\%} and \textbf{40.5\%} over training from scratch in simulation and real-world experiments, respectively. The results validate our core hypothesis that human manipulation data can serve as a powerful foundation for learning bimanual robotic manipulation policies. See our \href{https://embodiedfoundation.github.io/hrdt}{project page} for code and pretrained models.
\end{abstract} 


\section{Introduction}

Recent advances in robotic learning have been driven by specialized action policies like ACT~\citep{zhao2023learning}, Diffusion Policy~\citep{chi2023diffusion}, and 3D Diffusion Policy~\citep{ze20243d}, as well as Vision-Language-Action (VLA) models such as RT-2~\citep{brohan2023rt}, OpenVLA~\citep{kim2024openvla}, RDT~\citep{liu2024rdt}, $\boldsymbol{\pi}_0$~\citep{black2024pi_0}, and $\boldsymbol{\pi}_{0.5}$~\citep{intelligence2025pi_}. However, these approaches face fundamental data collection challenges. Robot demonstration data relies heavily on teleoperation~\citep{zhao2023learning, aldaco2024aloha}, which requires expensive equipment and skilled operators, while advanced data collection systems like Universal Manipulation Interface~\citep{chi2024universal} and motion capture setups~\citep{wang2024dexcap, xu2025dexumi} suffer from complex infrastructure requirements and inconsistent data quality that limit scalability.

Current VLA models typically employ cross-embodiment pre-training on robot datasets like Open X-Embodiment~\citep{o2024open} and AgiBot World Colosseo~\citep{bu2025agibot}. This approach faces two critical limitations: the diverse morphologies and action spaces across robot embodiments make unified training challenging, and existing robot datasets remain limited in scale with heterogeneous data quality across different collection setups. These constraints fundamentally limit the data availability and generalization capabilities needed for general-purpose robotic manipulation~\citep{team2024octo, o2024open}.

In stark contrast, human manipulation behaviors represent a vast, readily accessible repository of demonstration data. The recent emergence of large-scale egocentric video datasets with detailed hand pose annotations, exemplified by EgoDex~\citep{hoque2025egodex} with its 829 hours of manipulation videos, offers unprecedented opportunities for learning rich behavioral priors. Human demonstrations naturally capture object affordances, manipulation strategies, and task decomposition patterns that could potentially serve as powerful inductive biases for robotic learning. Recent works have begun exploring this direction: EgoMimic~\citep{kareer2024egomimic} employs co-training on human and robot data with egocentric video, while Humanoid Policy (HAT)~\citep{qiu2025humanoid} uses differentiable retargeting for human-humanoid behavior modeling.

This paper introduces H-RDT (Human to Robotics Diffusion Transformer), a novel approach that systematically leverages large-scale egocentric human manipulation data to enhance robot manipulation capabilities. Our approach focuses on three specific aspects:
\textbf{Data Scarcity:} We harness the abundance of human manipulation videos with 3D hand pose annotations to provide rich behavioral priors that capture natural manipulation strategies, object affordances, and task decomposition patterns.
\textbf{Cross-Embodiment Transfer:} We develop a modular transformer architecture with specialized action encoders and decoders that enable effective knowledge transfer from human demonstrations to diverse robotic platforms while preserving learned manipulation knowledge.
\textbf{Training Efficiency:} We employ a two-stage training paradigm with flow matching, first pre-training on large-scale human data followed by cross-embodiment fine-tuning, enabling stable and efficient policy learning throughout.

Our method introduces structural and training method innovations for human-to-robot knowledge transfer through human manipulation data pre-training.

The main contributions of this work are:
\begin{itemize}
\item A novel framework for systematically leveraging large-scale egocentric human manipulation data to enhance robotic policy learning
\item A diffusion transformer architecture with modular human-to-robot transfer components that enables effective cross-embodiment knowledge transfer
\item A comprehensive empirical validation demonstrating consistent improvements over state-of-the-art methods across simulation and real-world scenarios
\item Insights into the value of human manipulation priors for sample-efficient robot learning, particularly in few-shot settings
\end{itemize}

\section{Related Work}

\subsection{Learning-based Robotic Manipulation}

Recent advances in imitation learning have been driven by specialized action policies including ACT~\citep{zhao2023learning}, Diffusion Policy~\citep{chi2023diffusion}, and 3D Diffusion Policy~\citep{ze20243d}. These action policies focus on learning direct visuomotor control for manipulation tasks, showing promising results on dexterous manipulation through advanced sequence modeling and generative approaches.

The emergence of Vision-Language-Action (VLA) models represents a significant paradigm shift toward more generalizable robotic systems. Recent VLA approaches include RT-2~\citep{brohan2023rt}, OpenVLA~\citep{kim2024openvla}, the Robotics Diffusion Transformer (RDT)~\citep{liu2024rdt}, $\boldsymbol{\pi}_0$~\citep{black2024pi_0}, $\boldsymbol{\pi}_{0.5}$~\citep{intelligence2025pi_}, and other VLA models~\citep{zhao2025cot, zhen20243d, liu2025hybridvla, wen2025tinyvla, wen2025dexvla}. These models combine visual understanding, language comprehension, and action generation within unified architectures, enabling instruction-following capabilities and cross-embodiment generalization through large-scale datasets~\citep{o2024open, wu2024robomind, khazatsky2024droid, fang2023rh20t}.

Our work builds upon the RDT architecture while introducing novel structural and training method innovations. Specifically, we adopt flow matching~\citep{lipman2022flow, liu2022rectified} as our training paradigm, which offers improved stability and efficiency compared to traditional diffusion training~\citep{esser2024scaling, bao2023all}. More importantly, we introduce novel human-to-robot knowledge transfer mechanisms that enable large-scale pre-training on human manipulation data followed by cross-embodiment fine-tuning.

\subsection{Learning from Egocentric Human Manipulation}
Large-scale egocentric datasets~\citep{grauman2022ego4d, damen2018scaling, liu2022hoi4d, banerjee2025hot3d, grauman2024ego} contain tens to hundreds of hours capturing human-object interaction, yet lack precise 3D hand-pose annotations required for dexterous manipulation learning. EgoDex~\citep{hoque2025egodex} addresses this gap by providing 829 hours (338k episodes) of egocentric video with per-frame 3D hand poses and language descriptions.

EgoMimic~\citep{kareer2024egomimic} and Humanoid Policy (HAT)~\citep{qiu2025humanoid} pioneer the use of egocentric human videos, yet both operate at modest scales: EgoMimic trains on 2k human demos, and HAT on 27k demos—orders of magnitude smaller than the 338k trajectories (829h) employed by H-RDT. Moreover, these works target a single humanoid embodiment; EgoMimic requires paired robot data during co-training, while HAT's retargeting assumes humanoid kinematics. H-RDT, in contrast, decouples large-scale human pre-training from robot-specific fine-tuning and generalizes to arbitrary robot morphologies via modular action adapters. Additional studies explore data augmentation techniques~\citep{li2025h2r} and paired human–robot data collection~\citep{xie2025human2robot}.

\section{Method}

In this section, we present H-RDT (Human to Robotics Diffusion Transformer), a novel approach for leveraging large-scale human manipulation data to enhance robotic policy learning.

\subsection{Problem Formulation}

%A key distinction of our approach lies in the action representation $\mathbf{a}_t$. Unlike methods that learn a latent action space from videos—which often require an auxiliary model and can yield ambiguous representations—we leverage explicit and structured 3D hand pose annotations from human demonstration data. This explicit representation provides a strong inductive bias for learning complex manipulation skills and facilitates more direct and effective knowledge transfer to robotic embodiments.

We formulate robotic manipulation as a conditional sequence generation problem where the goal is to learn a policy $\pi_\theta$ that generates action sequences $\mathbf{a}_{t:t+H} = \{\mathbf{a}_t, \mathbf{a}_{t+1}, \ldots, \mathbf{a}_{t+H-1}\}$ given multimodal observations. Formally, at each timestep $t$, the agent observes visual observations $\mathbf{o}_t \in \mathbb{R}^{H \times W \times 3}$ from one or more RGB cameras, proprioceptive state $\mathbf{s}_t \in \mathbb{R}^{d_s}$ encoding current robot state and gripper state, and language instruction $\mathbf{l} \in \mathbb{R}^{L \times d_\text{lang}}$ describing the task. The policy outputs a sequence of future actions $\mathbf{a}_{t:t+H}$ where each action $\mathbf{a}_i \in \mathbb{R}^{d_a}$ represents robot control commands (e.g., joint positions, end-effector poses) over a prediction horizon $H$.

To achieve a generalist policy, large-scale imitation learning is required, while the data for a specified embodiment are scarce. To counter this, current methods majorly turn to train with demonstrations from multiple heterogeneous embodiments~\cite{liu2024rdt,black2024pi_0,intelligence2025pi_}. However, the total scale of the data remains limited due to the high cost of teleoperation. 

An alternative approach is to leverage egocentric human manipulation data, which could potentially provide data from a unified human embodiment with manipulation priors transferable across diverse robot platforms, thereby reducing the conflicts of learning from heterogeneous embodiments while enabling low-cost data acquisition. However, this approach faces three main challenges: Firstly, existing methods operate at modest scales with limited human manipulation data, failing to fully exploit the potential of human behavioral priors for robotic learning. Secondly, the significant embodiment differences between humans and robots, including end effector types and forward kinematics~\cite{qiu2025humanoid}, make it challenging to effectively transfer manipulation knowledge from human demonstrations to target robots. Thirdly, while concurrent work enables manipulation on specific robots using paired human data~\cite{kareer2024egomimic,qiu2025humanoid}, it remains largely under-addressed how to build a foundation model that can be efficiently adapted to multiple diverse robot embodiments through fine-tuning on robot-specific data.

% Figure 2: H-RDT Architecture Overview 
\begin{figure*}[!htbp]
\centering
\includegraphics[width=0.75\textwidth]{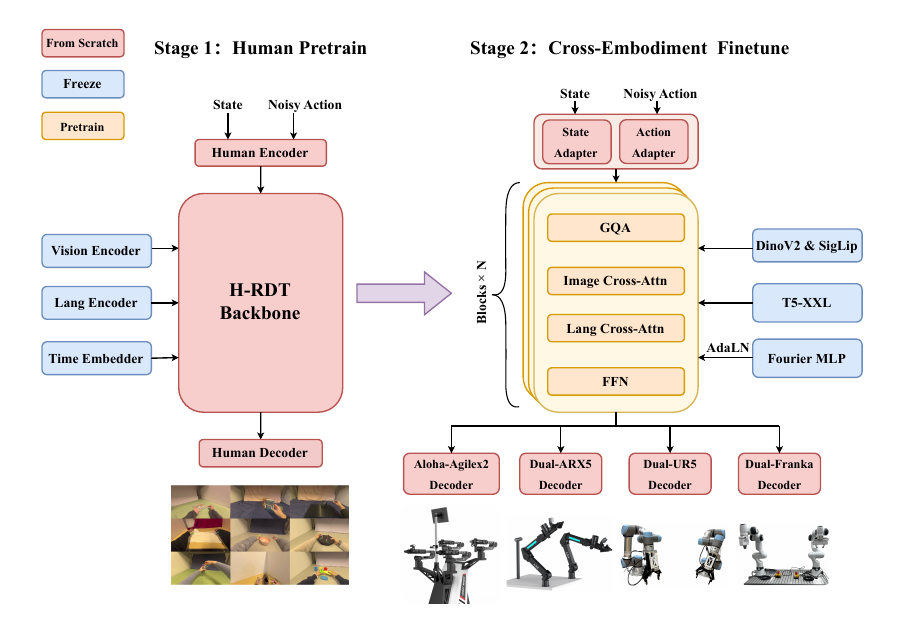}
\caption{\textbf{H-RDT framework.} Our approach consists of two main stages: (1) pre-training on large-scale human manipulation data with 48-dimensional hand pose representations, and (2) cross-embodiment fine-tuning with modular action encoders and decoders adapted to specific robot action spaces.}
\label{fig:overview}
\end{figure*}

\subsection{Overview}

To address the aforementioned challenges, we propose H-RDT (Human to Robotics Diffusion Transformer), as illustrated in Figure~\ref{fig:overview}, a transformer-based architecture trained with a structured paradigm to learn from human data. To counter the embodiment mismatch between humans and robots, H-RDT builds upon a shared action representation space to bridge human and robot embodiments, and satisfies the need for scalable cross-embodiment deployment by utilizing a two-stage training paradigm. Finally, H-RDT leverages flow matching and a scalable Transformer-based architecture for stable and expressive policy learning.

\subsection{Human Action Representation Design}

To address the challenge of embodiment differences between humans and robots, current methods either use flow as a transit representation for action~\citep{xu2024flow, wen2024any}, which provides only high-level object motion guidance without explicit action parameters and requires additional policy networks to translate flow into robot-specific controls, or require detailed re-targeting between human pose and target robot, which constrains the applicability of learning policy. To this end, we utilize detailed 3D hand poses, where actions are represented as compact 48-dimensional vectors capturing essential bimanual dexterous information:

\begin{itemize}
\item Bilateral wrist poses (position (3D) and orientation (6D) for both hands): 18 dimensions, which are identical to the End-Effector pose of robots
\item Fingertip positions (3D coordinates for all fingers on both hands): 30 dimensions
\end{itemize}

This representation serves as a superset for the action space of most current robots controlled with the End-Effector poses, thereby ensuring effective knowledge distillation across distinct kinematic structures. This structured encoding explicitly represents fundamental manipulation dynamics and spatial relationships crucial for generalizable manipulation, effectively mitigating embodiment discrepancies by focusing on universally transferable features such as grasp configurations, orientation constraints, and relative positional dynamics.

\subsection{Two-Stage Training Paradigm}

Concurrent work that learns robot policy from human data mostly requires rigorous pairing relationship between human and target embodiment, thereby failing to adapt to multiple embodiments during deployment. To solve this issue, we adopt a carefully designed two-stage training paradigm that maximizes the benefit of human demonstration data while enabling effective cross-embodiment robot deployment. Rather than a traditional diffusion objective, H-RDT employs flow matching~\citep{lipman2022flow} for action generation, offering superior training stability and inference efficiency.

\paragraph{Stage 1: Human Data Pre-training}
In the first stage, we train H-RDT on the complete EgoDex dataset with 48-dimensional human hand action representations. Concretely, our model is trained on 338K+ trajectories across 194 distinct manipulation tasks using the complete EgoDex dataset~\cite{hoque2025egodex}, providing comprehensive coverage of human manipulation strategies, object interactions, and bimanual coordination patterns. 
% The pre-training is conducted on 8 H100 GPUs for 500K steps.
% \item \textbf{Large-Scale Training:} We employ multi-nodes for distributed training over 500K steps, enabling the model to fully capture the rich diversity of human manipulation behaviors from the extensive egocentric video data
% \item \textbf{Vision-Language Grounding:} The dual cross-attention mechanism processes both visual observations and natural language instructions, learning to ground language descriptions with visual manipulation patterns
% \item \textbf{Flow Matching Training:} The model learns to generate human-like manipulation trajectories using flow matching objectives, developing robust action priors for dexterous bimanual manipulation that transfer effectively to robotic systems
% \end{itemize}

\paragraph{Stage 2: Cross-Embodiment Fine-tuning}
To quickly adapt pre-trained for cross-embodiment deployment, the second stage adapts the pre-trained model to specific robot embodiments through selective weight transfer and modular re-initialization:  The vision encoder, language encoder, and transformer backbone weights are transferred from the pre-trained model, preserving learned multi-modal representations and manipulation priors developed from human demonstrations.
The state adaptor (MLP$_{\text{state}}$), action adaptor (MLP$_{\text{action}}$), and action decoder are completely reinitialized to handle the target robot's action space (e.g., 14 dimensions for dual 7-DOF arms with parallel grippers)

This selective transfer strategy ensures that learned manipulation semantics from human demonstrations are preserved while enabling adaptation to diverse robot morphologies. The modular design allows action encoders and decoders to be retrained from scratch for each target embodiment without compromising the learned visual-semantic representations. 

\subsection{H-RDT Architecture}

\subsubsection{Flow Matching for Action Generation}

Rather than traditional diffusion training, H-RDT employs flow matching~\citep{lipman2022flow} for action generation, offering superior training stability and inference efficiency compared to traditional diffusion modeling. Flow matching learns a vector field that transforms a simple noise distribution to the target action distribution through a continuous normalizing flow.

Given a target action sequence $\mathbf{a}^*_{t:t+H}$, we construct a straight-line flow path:
\begin{align}
\mathbf{a}_\tau = \tau \cdot \mathbf{a}^*_{t:t+H} + (1-\tau) \mathbf{z}
\end{align}
where $\mathbf{z} \sim \mathcal{N}(0, \mathbf{I})$ is Gaussian noise and $\tau \in [0,1]$ parameterizes the flow time. The neural network $v_\theta$ learns to predict the vector field:
\begin{align}
\mathcal{L}_{\text{FM}} = \mathbb{E}_{\tau, \mathbf{z}, \mathbf{a}^{*}, \mathbf{c}} \left[ \| v_\theta(\mathbf{a}_\tau, \tau, \mathbf{c}) - (\mathbf{z} - \mathbf{a}^*_{t:t+H}) \|^2 \right]
\end{align}
where $\mathbf{c} = \{\mathbf{o}_t, \mathbf{s}_t, \mathbf{l}\}$ represents the conditioning information including multi-view RGB observations $\mathbf{o}_t$, proprioception $\mathbf{s}_t$, and language instruction $\mathbf{l}$. During inference, we sample actions by integrating the learned vector field using an ODE solver with deterministic steps (detailed implementation in Appendix B.3).

\subsubsection{Network Architecture}

H-RDT adopts a unified transformer architecture comprising five modular components: a vision encoder, language encoder, modular action encoder, transformer backbone, and modular action decoder. 

\textbf{Vision and Language Encoders:} RGB observations are encoded using pre-trained visual backbones DinoV2~\citep{oquab2023dinov2} and SigLIP~\citep{zhai2023sigmoid}, followed by MLP adapters that project image features into the embedding space of dimension $d_{\text{model}}$. Text instructions are embedded using a pre-trained T5-XXL language model~\citep{raffel2020exploring} and projected via similar adapters.

\textbf{Modular Action Encoder:} The proprioceptive state $\mathbf{s}_t$ and noisy action sequence $\mathbf{a}^\tau_{t: t+H}$ are encoded with modular MLP adapters:
\begin{align}
\mathbf{h}_{\text{state}} &= \text{StateAdapter}(\mathbf{s}_t) \in \mathbb{R}^{d_{\text{model}}}, \\
\mathbf{h}_{\text{action}} &= \text{ActionAdapter}(\mathbf{a}^\tau_{t: t+H}) \in \mathbb{R}^{H \times d_{\text{model}}},
\end{align}
where $\mathbf{a}^\tau_{t: t+H}$ represents the noisy action sequence at flow time $\tau$ used in flow matching training, and $H$ denotes the prediction horizon.

\textbf{Transformer Backbone:} H-RDT adopts the LLaMA-3 architecture style~\citep{touvron2023llama} with RMSNorm layer normalization and SwiGLU activation functions. Each transformer block processes the concatenated input $\mathbf{x} = \text{Concat}(\mathbf{h}_{\text{state}}, \mathbf{h}_{\text{action}})$ using self-attention, while image and language features are injected via separate cross-attention to avoid modality imbalance~\cite{liu2024rdt}. The flow time $\tau$ is mapped into timestep embeddings and integrated via AdaLN~\citep{peebles2023scalable}.

\textbf{Modular Action Decoder:} The predicted hidden states $\mathbf{h}_{\text{action}}$ are decoded using a modular MLP:
\begin{align}
\hat{\mathbf{a}}_{t:t+H} = \text{ActionDecoder}(\mathbf{h}_{\text{action}}, \mathbf{t}_{\text{emb}}).
\end{align}
where $\mathbf{t}_{\text{emb}}$ represents timestep embeddings for flow matching, and the decoder outputs actions in the target robot's action space. The modular action encoder and decoder are re-initialized during cross-embodiment fine-tuning.

\section{Experiments}

\subsection{Experimental Setup}

We conduct comprehensive experiments to evaluate H-RDT's effectiveness across simulation and real-world scenarios. Our evaluation covers four key dimensions: (1) single-task and multi-task performance across diverse manipulation scenarios, (2) cross-embodiment generalization across diverse robotic platforms, (3) environmental robustness through domain randomization, and (4) sample efficiency in few-shot learning with limited real-world demonstrations.

\textbf{Simulation Environment:} We use the RoboTwin 2.0 platform~\citep{chen2025robotwin}, a comprehensive dual-arm manipulation benchmark featuring diverse household tasks. The platform provides two evaluation modes: Easy mode with clean tabletop environments, and Hard mode with domain randomization including 3cm table height variation, random backgrounds, lighting changes, and object clutter.

\textbf{Robot Embodiment:} To demonstrate cross-embodiment transfer capabilities, we evaluate H-RDT across multiple robotic platforms in both simulation and real-world settings. Simulation experiments cover two distinct embodiments: Aloha-Agilex-1.0 and dual-arm Franka-Panda. Real-world validation uses three different platforms: dual-arm ARX5, Aloha-Agilex-2.0 (dual-arm Piper), and UR5 + UMI configuration.

\textbf{Training Configuration:} Detailed training configurations for different experimental settings are provided in Appendix C.

\subsection{Baselines and Comparison Methods}

We compare H-RDT against several state-of-the-art methods:
\begin{itemize}
\item \textbf{RDT}~\citep{liu2024rdt}: Robotics Diffusion Transformer baseline
\item \textbf{$\boldsymbol{\pi}_0$}~\citep{black2024pi_0}: State-of-the-art vision-language-action model
\item \textbf{w/o human}: Our model without human pre-training
\end{itemize}

\subsection{Real-world Validation}

We evaluate H-RDT across three distinct real-world robotic platforms to validate cross-embodiment transfer capabilities and robustness in practical deployment scenarios. All real-world experiments employ multi-task training.

\subsubsection{Aloha-Agilex-2.0 Experiments}

We evaluate H-RDT on the Aloha-Agilex-2.0 platform (dual-arm Piper) across two bimanual manipulation tasks.

\paragraph{Task 1: Towel Folding} This deformable object manipulation task tests the model's ability to handle non-rigid materials through sequential folding operations. 

\paragraph{Task 2: Cup to Coaster Placement} This spatial reasoning task requires the model to automatically select the appropriate hand based on object location: cups on the left side must be grasped with the left hand, while cups on the right side must be grasped with the right hand.

Both tasks employ sub-task-based scoring systems that evaluate progressive completion levels, with final evaluation focusing on complete success rates. Tables~\ref{tab:towel_folding} and~\ref{tab:cup_coaster} present the performance breakdown across methods, with each task evaluated over 25 trials.

\begin{table}[!t]
\centering
\small
\begin{tabular}{l|ccc}
\hline
Performance Level & RDT & w/o human & H-RDT \\
\hline
0.0: Complete failure & - & 7 & - \\
\hline
0.2: One fold single side & - & 18 & - \\
\hline
0.5: One fold both sides & 3 & - & 3 \\
\hline
0.7: Two fold incomplete & 12 & - & 9 \\
\hline
1.0: Two fold complete & 10 & - & 13 \\
\hline
Complete success rate & 40\% & 0\% & \textbf{52\%} \\
\hline
\end{tabular}
\caption{Towel folding task results on Aloha-Agilex-2.0 platform with detailed performance breakdown.}
\label{tab:towel_folding}
\end{table}

\begin{table}[!t]
\centering
\small
\begin{tabular}{l|ccc}
\hline
Performance Level & RDT & w/o human & H-RDT \\
\hline
0.0: Complete failure & 10 & 14 & 6 \\
\hline
0.4: Grasped, failed to place & 8 & 6 & 3 \\
\hline
1.0: Successfully completed & 7 & 5 & 16 \\
\hline
Complete success rate & 28\% & 20\% & \textbf{64\%} \\
\hline
\end{tabular}
\caption{Cup to coaster placement task results on Aloha-Agilex-2.0 platform with detailed performance breakdown.}
\label{tab:cup_coaster}
\end{table}

For the towel folding task (Table~\ref{tab:towel_folding}), H-RDT achieves a 52\% complete success rate compared to 40\% for RDT and 0\% for training from scratch. The model without human data fails to achieve any complete folding, showing only partial success at lower skill levels, while RDT and H-RDT demonstrate more sophisticated manipulation capabilities.

The cup-to-coaster task (Table~\ref{tab:cup_coaster}) shows H-RDT achieving a 64\% complete success rate compared to 28\% for RDT and 20\% for training without human data. H-RDT demonstrates the lowest failure rate and fewer instances of partial success, indicating more robust performance for tasks requiring spatial reasoning to select appropriate arms.

Overall, H-RDT achieves an average success rate of 58\% across both bimanual tasks compared to 34\% for RDT and 10\% for training without human data, demonstrating the effectiveness of human manipulation priors for diverse coordination challenges including deformable object manipulation and spatial reasoning tasks.

\subsubsection{Dual-arm ARX5 Few-shot Experiments}

To thoroughly validate the advantages of human manipulation priors, we design a challenging real-world experiment with a combination of massive task diversity and data scarcity: 113 diverse pick-and-place tasks using a dual-arm ARX5 robotic system, with only 1-5 demonstrations per task. This multi-task few-shot setting is specifically designed to test the limits of sample efficiency and highlight the value of human behavioral priors.

The EgoDex pretraining dataset contains extensive pick-and-place manipulation patterns similar to these tasks, providing rich prior knowledge about how to perform such operations. Under these challenging conditions—where even state-of-the-art models like $\boldsymbol{\pi}_0$ struggle to properly fit the limited demonstration trajectories—H-RDT's human manipulation priors enable noticeable performance improvements. H-RDT achieves an average success rate of 41.6\% compared to 16.0\% for RDT, 31.2\% for $\boldsymbol{\pi}_0$, and 17.6\% for H-RDT w/o human, demonstrating the value of human manipulation priors for few-shot learning in data-limited scenarios.

\begin{table*}[!t]
\centering
\small
\begin{tabular}{p{6cm}|*{4}{c}}
\hline
Task Category & RDT & $\boldsymbol{\pi}_0$ & w/o human & H-RDT \\
\hline
Pick yellow item to the plate & 12\% & 16\% & 28\% & \textbf{40\%} \\
Place banana and carrot in basket & 24\% & 40\% & 8\% & \textbf{44\%} \\
Stack two bowls together & 32\% & 60\% & 40\% & \textbf{68\%} \\
Place cube in front of highest chips & 12\% & 0\% & 0\% & \textbf{20\%} \\
Stack one can on top of the other can & 0\% & \textbf{40\%} & 12\% & 36\% \\
\hline
Average & 16.0\% & 31.2\% & 17.6\% & \textbf{41.6\%} \\
\hline
\end{tabular}
\caption{Real-world few-shot learning results on manipulation tasks. Each task has 1-5 demonstrations for training. H-RDT demonstrates superior sample efficiency and real-world transfer capabilities.}
\label{tab:real_world}
\end{table*}

\subsubsection{Dual UR5 + UMI Experiments}

We evaluate H-RDT on a dual UR5 robotic system with demonstrations collected using Universal Manipulation Interface (UMI)~\citep{chi2024universal}, a data collection framework that enables portable, low-cost human demonstration collection through hand-held grippers.

The evaluation focuses on bimanual takeout bag placement tasks decomposed into four sequential subtasks: right-hand pick, right-hand place, left-hand pick, and left-hand place.

Table~\ref{tab:ur5_umi} presents the success rates for each subtask across different methods, with each evaluation conducted over 25 trials.

\begin{table}[!t]
\centering
\small
\begin{tabular}{l|cccc}
\hline
Subtask & RDT & $\boldsymbol{\pi}_0$ & w/o human & H-RDT \\
\hline
Right hand pick & 36\% & 40\% & 20\% & \textbf{64\%} \\
Right hand place & 32\% & 28\% & 16\% & \textbf{56\%} \\
Left hand pick & 28\% & 36\% & 20\% & \textbf{60\%} \\
Left hand place & 20\% & 20\% & 8\% & \textbf{52\%} \\
\hline
Average & 29.0\% & 31.0\% & 16.0\% & \textbf{58.0\%} \\
\hline
\end{tabular}
\caption{Performance breakdown on dual UR5 + UMI takeout bag placement task showing success rates for individual subtasks.}
\label{tab:ur5_umi}
\end{table}

H-RDT achieves consistently superior performance across all subtasks, with an average success rate of 58.0\% compared to 29.0\% for RDT, 31.0\% for $\boldsymbol{\pi}_0$, and 16.0\% for training from scratch. The results show notable improvements in pick operations (64\% and 60\% for right and left hands, respectively) and 27-42\% absolute improvements over baseline methods, demonstrating the value of human manipulation priors for bimanual coordination.

\subsection{Simulation Results on RoboTwin 2.0}

\subsubsection{Single-Task Performance}

We evaluate single-task performance on 13 representative manipulation tasks from the RoboTwin 2.0 benchmark. Each task is trained on 50 demonstrations collected in clean environments and evaluated in two modes: Easy mode (clean tabletop environments matching training conditions) and Hard mode (challenging environments with domain randomization including lighting changes, object clutter, and table height variations). Detailed results for all tasks are provided in Table~\ref{tab:single_task} in the Appendix.

H-RDT achieves the highest average success rate of 68.7\% in Easy mode and 25.6\% in Hard mode, demonstrating significant improvements over existing methods. H-RDT substantially surpasses training from scratch (w/o human) by 8.4\% in both Easy and Hard modes, validating the effectiveness of human manipulation pre-training.

\subsubsection{Multi-Task Performance}

We conduct multi-task experiments on 45 tasks from RoboTwin 2.0, training on approximately 2250 demonstrations collected under domain randomization (Hard mode data). Table~\ref{tab:multi_task} shows the results on a representative subset of 10 tasks evaluated in Hard mode.

\begin{table}[!t]
\centering
\small
\begin{tabular}{l|*{4}{c}}
\hline
Task & RDT & $\boldsymbol{\pi}_0$ & w/o human & H-RDT \\
\hline
Click Alarmclock & 30\% & 69\% & 69\% & \textbf{94\%} \\
Click Bell & 13\% & 40\% & 77\% & \textbf{98\%} \\
Dump Bin Bigbin & 22\% & 33\% & 58\% & \textbf{89\%} \\
Grab Roller & 57\% & 71\% & 91\% & \textbf{97\%} \\
Handover Mic & 81\% & 97\% & 93\% & \textbf{99\%} \\
Move Playingcard Away & 3\% & 23\% & 44\% & \textbf{67\%} \\
Open Laptop & 33\% & 35\% & 88\% & \textbf{92\%} \\
Open Microwave & 19\% & 46\% & 32\% & \textbf{82\%} \\
Press Stapler & 15\% & 59\% & 71\% & \textbf{81\%} \\
Stack Bowls Three & 15\% & 11\% & 49\% & \textbf{73\%} \\
\hline
Average & 28.8\% & 48.4\% & 67.2\% & \textbf{87.2\%} \\
\hline
\end{tabular}
\caption{Multi-task success rates (\%) on RoboTwin 2.0 benchmark (Hard mode evaluation). Results show performance after multi-task training on 50 tasks with domain randomized data.}
\label{tab:multi_task}
\end{table}

In multi-task settings, H-RDT achieves an average success rate of 87.2\%, significantly outperforming RDT (28.8\%), $\boldsymbol{\pi}_0$ (48.4\%), and H-RDT w/o human (67.2\%). H-RDT shows a substantial 20.0\% absolute improvement over training from scratch, which is notably larger than the improvements observed in single-task scenarios, demonstrating that human manipulation pre-training provides even greater advantages when learning across diverse tasks simultaneously.

\subsubsection{Cross-Embodiment Generalization}

\begin{figure}[t!]
\centering
\includegraphics[width=0.48\textwidth]{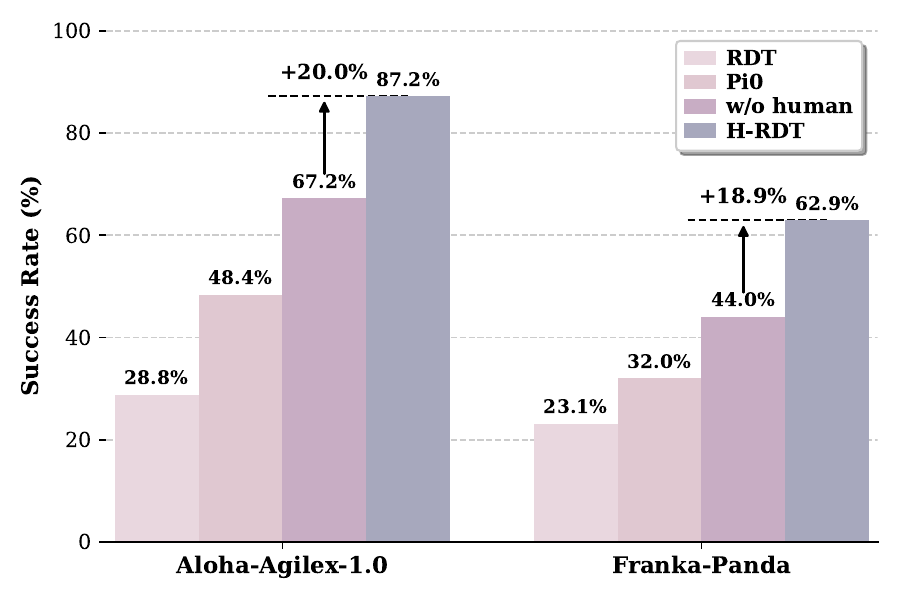}
\caption{Cross-embodiment multi-task performance on RoboTwin 2.0 tasks.}
\label{fig:cross_embodiment}
\end{figure}

To further validate the cross-embodiment transfer capabilities of H-RDT, we conduct multi-task experiments across two different robotic embodiments in simulation. We evaluate both Aloha-Agilex-1.0 and Franka-Panda platforms using the same experimental setup as described above. Figure~\ref{fig:cross_embodiment} shows the performance comparison across these platforms.

H-RDT demonstrates strong performance across both embodiments, achieving 87.2\% on Aloha-Agilex-1.0 and 62.9\% on Franka-Panda, significantly outperforming baseline methods on both platforms. Detailed per-task results for Franka-Panda are provided in Table~\ref{tab:franka_detailed} in the Appendix. The consistent improvements across different robotic morphologies validate the cross-embodiment generalization capabilities of our modular action encoder design.

\subsection{Analysis and Discussion}

\noindent\textbf{Impact of Human Pre-training:} The consistent improvements of H-RDT over the baseline without human data across all experimental settings validate our core hypothesis that human manipulation data provides valuable inductive biases. The benefits are most pronounced in few-shot real-world scenarios, where human priors about object affordances and manipulation strategies prove crucial.

\noindent\textbf{Environmental Robustness:} H-RDT demonstrates strong performance under challenging conditions in RoboTwin 2.0 Hard mode with domain randomization. The model successfully handles environmental variations including lighting changes, object clutter, and table height variations, consistently outperforming baseline methods.

\noindent\textbf{Sample Efficiency:} In few-shot real-world experiments, H-RDT's ability to learn from just 1-5 demonstrations per task significantly outperforms baselines, highlighting the practical value of human behavior priors for reducing data requirements in robotic learning.

\noindent\textbf{Task Diversity and Complexity:} Real-world experiments demonstrate H-RDT's capability to handle diverse manipulation challenges, including deformable object manipulation and tasks requiring spatial reasoning, showcasing its versatility across different manipulation complexities.

\noindent\textbf{Cross-Platform Robustness:} Our comprehensive evaluation across both simulation and real-world settings demonstrates H-RDT's robust performance across diverse robotic embodiments, including Aloha-Agilex-1.0, dual-arm Piper, dual-arm ARX5, dual-arm Franka-Panda, and dual-arm UR5+UMI platforms. This cross-platform consistency validates the effectiveness of our modular architecture design and human-to-robot knowledge transfer approach.

\section{Conclusion}

This paper introduces H-RDT, a novel approach that leverages large-scale egocentric human manipulation videos with 3D hand pose annotations to enhance robotic manipulation capabilities. Our central contribution shows that rich manipulation knowledge can be learned from human behavioral priors and adapted to diverse robotic manipulation tasks.

The key technical innovations include: (1) a modular transformer architecture with specialized action encoders and decoders that enable cross-embodiment adaptation, (2) flow matching for stable and efficient policy learning, and (3) a two-stage training paradigm that first pre-trains on human data before fine-tuning on robot-specific data.

Comprehensive evaluations demonstrate consistent improvements over state-of-the-art methods, validating that human manipulation priors provide powerful inductive biases for sample-efficient robotic learning.

\bibliography{main}

% Add appendix content
\clearpage
\appendix

\section{A. Additional Experimental Results}

\subsection{A.1 Single-Task Performance Results (RoboTwin 2.0 Benchmark)}

This section provides comprehensive per-task results for 13 manipulation tasks evaluated in the single-task experiments on the RoboTwin 2.0 benchmark. Table~\ref{tab:single_task} presents the detailed success rates across all baseline methods in both Easy and Hard evaluation modes.

To accelerate training speed in single-task experiments, we concatenate three 240×320 views into a single 360×320 input for training, which may result in some performance degradation compared to higher resolution settings.

\subsection{A.2 Franka-Panda Detailed Results}

This section provides comprehensive per-task results for the Franka-Panda embodiment on the 10-task subset used for detailed evaluation in the main paper.

\begin{table}[!htbp]
\centering
\small
\begin{tabular}{l|*{4}{c}}
\hline
Task & RDT & $\boldsymbol{\pi}_0$ & w/o human & H-RDT \\
\hline
Click Alarmclock & 58\% & 71\% & 59\% & \textbf{94\%} \\
Click Bell & 18\% & 48\% & 39\% & \textbf{89\%} \\
Dump Bin Bigbin & 8\% & 13\% & 21\% & \textbf{31\%} \\
Grab Roller & 50\% & 35\% & 70\% & \textbf{75\%} \\
Handover Mic & 13\% & 17\% & 53\% & \textbf{64\%} \\
Move Playingcard Away & 3\% & 7\% & 17\% & \textbf{20\%} \\
Open Laptop & 15\% & 11\% & 41\% & \textbf{43\%} \\
Open Microwave & 1\% & 45\% & 29\% & \textbf{59\%} \\
Press Stapler & 34\% & 42\% & 51\% & \textbf{86\%} \\
Shake Bottle & 31\% & 31\% & 60\% & \textbf{68\%} \\
\hline
Average & 23.1\% & 32.0\% & 44.0\% & \textbf{62.9\%} \\
\hline
\end{tabular}
\caption{Detailed Franka-Panda multi-task results (\%) on 10-task subset.}
\label{tab:franka_detailed}
\end{table}

\section{B. Implementation Details}

\subsection{B.1 Model Architecture}

Table~\ref{tab:architecture_details} provides the key hyperparameter settings for the H-RDT model architecture.

\begin{table}[!htbp]
\centering
\small
\begin{tabular}{l|c}
\hline
\textbf{Component} & \textbf{Configuration} \\
\hline
\multicolumn{2}{l}{\textbf{Transformer Backbone}} \\
Hidden Size & 2176 \\
Layers & 16 \\
Attention Heads & 16 \\
Key-Value Heads (GQA) & 8 \\
Layer Norm Epsilon & 1e-5 \\
Activation Function & SwiGLU \\
\hline
\multicolumn{2}{l}{\textbf{Action Dimensions}} \\
Human Hand Pose (Pretrain) & 48 \\
Dual 6-DOF Arms (Finetune) & 14 \\
\hline
\multicolumn{2}{l}{\textbf{Camera Views}} \\
Pretrain & 1 (egocentric) \\
Finetune & 1-3 (robot-dependent) \\
\hline
\multicolumn{2}{l}{\textbf{Adapter Networks}} \\
State/Action Adaptor & 3-layer MLP + SiLU \\
Image/Language Adaptor & 2-layer MLP + SiLU \\
\hline
\multicolumn{2}{l}{\textbf{Flow Matching}} \\
Inference Steps & 5 \\
Timestep Range & [0, 0.999] \\
Sampling Strategy & Uniform \\
\hline
\multicolumn{2}{l}{\textbf{Model Scale}} \\
Total Parameters & ~2B \\
\hline
\end{tabular}
\caption{H-RDT architecture hyperparameters and key configuration settings.}
\label{tab:architecture_details}
\end{table}

\subsection{B.2 Training Configuration and Data Processing}

\textbf{Training Configuration:} Both pre-training and fine-tuning use AdamW optimizer with learning rate 1e-4, weight decay 0.01, and gradient clipping at norm 1.0. We employ mixed-precision training (bfloat16) for computational efficiency and gradient accumulation to maintain effective batch sizes.

\noindent\textbf{Data Processing:} Images are processed with 196 patches for vision encoding. Language instructions are tokenized with a maximum length of 1024 tokens. During pre-training, we process the full 338K+ trajectory EgoDex dataset with 48-dimensional human hand actions. Fine-tuning adapts to target robot action spaces.

\subsection{B.3 Details on Flow Matching}

\noindent\textbf{Training Implementation:} We employ uniform timestep sampling with $\tau \in [0, 0.999]$ during training.

\noindent\textbf{Inference Implementation:} During inference, we start with Gaussian noise $\mathbf{a}_0 \sim \mathcal{N}(0, \mathbf{I})$ and integrate the learned vector field using a deterministic ODE solver with 5 function evaluations and step size $\Delta t = 1/5 = 0.2$. At each step, we update the action as $\mathbf{a}_{t+\Delta t} = \mathbf{a}_t + \Delta t \cdot v_\theta(\mathbf{a}_t, t, \mathbf{c})$ for real-time performance (30Hz control frequency).

\subsection{B.4 Real-World Task Definitions}

\begin{figure*}[t!]
\centering
\includegraphics[width=0.9\textwidth]{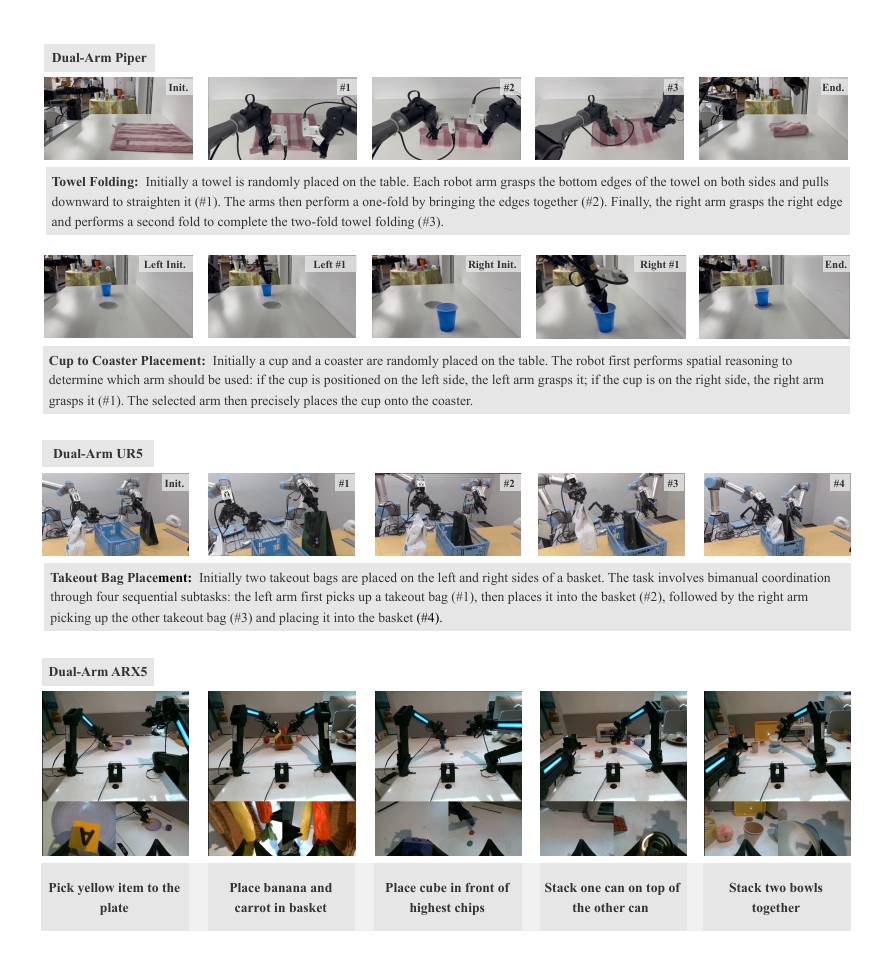}
\caption{Task definition of real-world experiments.}
\label{fig:task_definitions}
\end{figure*}

Figure~\ref{fig:task_definitions} provides visual illustrations of the real-world manipulation task in our experiments.

\begin{table*}[!htbp]
\centering
\footnotesize
\begin{tabular}{l|cc|cc|cc|cc|cc|cc}
\hline
\multirow{2}{*}{Task} & \multicolumn{2}{c|}{ACT} & \multicolumn{2}{c|}{DP} & \multicolumn{2}{c|}{RDT} & \multicolumn{2}{c|}{$\boldsymbol{\pi}_0$} & \multicolumn{2}{c|}{w/o human} & \multicolumn{2}{c}{H-RDT} \\
& Easy & Hard & Easy & Hard & Easy & Hard & Easy & Hard & Easy & Hard & Easy & Hard \\
\hline
Grab Roller & 66\% & 6\% & 98\% & 1\% & 74\% & 43\% & 96\% & \textbf{80\%} & \textbf{99\%} & 37\% & 95\% & 63\% \\
Handover Mic & 9\% & 0\% & 53\% & 0\% & 98\% & 41\% & \textbf{100\%} & 13\% & \textbf{100\%} & 4\% & \textbf{100\%} & \textbf{14\%} \\
Lift Pot & 7\% & 2\% & 37\% & 0\% & 82\% & 19\% & 84\% & \textbf{36\%} & \textbf{97\%} & 11\% & 94\% & 27\% \\
Move Can Pot & 0\% & 0\% & 42\% & 0\% & 47\% & 23\% & \textbf{74\%} & \textbf{32\%} & 63\% & 5\% & 54\% & 26\% \\
Open Laptop & 32\% & 0\% & 46\% & 0\% & 61\% & 36\% & 85\% & 46\% & 91\% & \textbf{59\%} & \textbf{92\%} & 42\% \\
Pick Dual Bottles & 4\% & 0\% & 26\% & 0\% & 42\% & 13\% & 57\% & 12\% & 37\% & 16\% & \textbf{67\%} & \textbf{20\%} \\
Place Object Basket & 0\% & 0\% & 17\% & 0\% & 42\% & 14\% & \textbf{62\%} & 10\% & 52\% & 15\% & \textbf{62\%} & \textbf{19\%} \\
Place Dual Shoes & 0\% & 0\% & 7\% & 0\% & 4\% & 4\% & 15\% & 0\% & 14\% & 4\% & \textbf{32\%} & \textbf{9\%} \\
Place Phone Stand & 0\% & 0\% & 17\% & 0\% & 15\% & 6\% & 35\% & 7\% & 25\% & 2\% & \textbf{36\%} & \textbf{9\%} \\
Put Bottles Dustbin & 0\% & 0\% & 23\% & 0\% & 21\% & 4\% & 54\% & 13\% & 53\% & 14\% & \textbf{64\%} & \textbf{16\%} \\
Put Object Cabinet & 4\% & 18\% & 50\% & 17\% & 30\% & 30\% & \textbf{69\%} & 29\% & 58\% & \textbf{41\%} & 63\% & 37\% \\
Stack Blocks Two & 0\% & 0\% & 6\% & 0\% & 32\% & 1\% & 40\% & 1\% & 28\% & \textbf{3\%} & \textbf{50\%} & \textbf{3\%} \\
Stack Bowls Two & 0\% & 0\% & 34\% & 0\% & 73\% & 21\% & 73\% & 41\% & 67\% & 13\% & \textbf{84\%} & \textbf{48\%} \\
\hline
Average & 9.4\% & 2.0\% & 35.1\% & 1.4\% & 47.8\% & 19.6\% & 64.9\% & 24.6\% & 60.3\% & 17.2\% & \textbf{68.7\%} & \textbf{25.6\%} \\
\hline
\end{tabular}
\caption{Single-task success rates (\%) on RoboTwin 2.0 benchmark. H-RDT achieves the highest average performance across both Easy and Hard evaluation modes. Bold indicates the best performance for each task and mode.}
\label{tab:single_task}
\end{table*}

\section{C. Training Configurations}

This section provides detailed training configurations for all experimental settings described in the main paper.

\subsection{C.1 Simulation Experiments}

\subsubsection{C.1.1 Single-Task Configuration}
\begin{itemize}
\item \textbf{Data:} 13 tasks, ~50 clean trajectories per task
\item \textbf{Platform:} Aloha-Agilex-1.0
\item \textbf{Training:} 10k steps, 4 H100 GPUs, batch size 16 per GPU
\item \textbf{Evaluation:} Easy mode (clean scenes) and Hard mode (domain randomization)
\end{itemize}

\subsubsection{C.1.2 Multi-Task Configuration}
\begin{itemize}
\item \textbf{Data:} 45 tasks, each 50 domain-randomized trajectories
\item \textbf{Platform:} Aloha-Agilex-1.0, Franka-Panda
\item \textbf{Training:} 30k steps for Aloha, 10k steps for Franka, 4 H100 GPUs, batch size 32 per GPU
\item \textbf{Evaluation:} Hard mode with domain randomization, three camera views
\end{itemize}

\subsection{C.2 Real-World Experiments}

\subsubsection{C.2.1 Dual-arm ARX5}
\begin{itemize}
\item \textbf{Data:} Few-shot pick-place tasks with 1-5 trajectories per task, 607 trajectories total across 113 tasks
\item \textbf{Platform:} Dual-arm ARX5
\item \textbf{Training:} 100k steps, 4 H100 GPUs
\end{itemize}

\subsubsection{C.2.2 Dual-arm UR5+UMI Platform}
\begin{itemize}
\item \textbf{Data:} Dual-hand takeout bag placement tasks, 100 trajectories
\item \textbf{Platform:} Dual-arm UR5+UMI Platform
\item \textbf{Training:} 20k steps, 8 H100 GPUs
\end{itemize}

\subsubsection{C.2.3 Aloha-Agilex-2.0 Platform}
\begin{itemize}
\item \textbf{Data:} Pouring water and folding clothes, each 50 trajectories
\item \textbf{Platform:} Aloha-Agilex-2.0 (dual-arm Piper)
\item \textbf{Training:} 50k steps, 8 H20 GPUs
\end{itemize}

\end{document}